# Quantum Artificial Intelligence for the Science of Climate Change

Manmeet Singh[1,2], Chirag Dhara[3], Adarsh Kumar[4], Sukhpal Singh Gill[5] and Steve Uhlig[5]

[1]Jackson School of Geosciences, University of Texas at Austin, Austin, USA
[2]Indian Institute of Tropical Meteorology, Ministry of Earth Sciences, India
[3]Krea University, Sri City, India
[4]Department of Systemics, School of Computer Science, University of Petroleum and Energy Studies, Dehradun, India
[5]School of Electronic Engineering and Computer Science, Queen Mary University of London, UK

manmeet.singh@utexas.edu, chirag.dhara.work@gmail.com, adarsh.kumar@ddn.upes.ac.in, s.s.gill@qmul.ac.uk, steve.uhlig@qmul.ac.uk

**Corresponding Author:** Manmeet Singh, Jackson School of Geosciences, University of Texas at Austin, Austin, TX 78712, United States, Email: manmeet.singh@utexas.edu

**Abstract**

Climate change has become one of the biggest global problems increasingly compromising the Earth's habitability. Recent developments such as the extraordinary heat waves in California & Canada, and the devastating floods in Germany point to the role of climate change in the ever increasing frequency of extreme weather. Numerical modelling of the weather and climate have seen tremendous improvements in the last five decades, yet stringent limitations remain to be overcome. Spatially and temporally localized forecasting is the need of the hour for effective adaptation measures towards minimizing the loss of life and property. Artificial Intelligence based methods are demonstrating promising results in improving predictions, but are still limited by the availability of requisite hardware and software required to process the vast deluge of data at a scale of the planet Earth. Quantum computing is an emerging paradigm that has found potential applicability in several fields. In this opinion piece, we argue that new developments in Artificial Intelligence algorithms designed for quantum computers - also known as Quantum Artificial Intelligence (QAI) - may provide the key breakthroughs necessary to furthering the science of climate change. The resultant improvements in weather and climate forecasts are expected to cascade to numerous societal benefits.

**1. Introduction**

The Earth's mean temperature has risen steeply over the last few decades precipitating a broad spectrum of global-scale impacts such as glacier melt, sea-level rise and an increasing frequency of weather extremes. These changes have resulted from the rising atmospheric carbon pollution during the industrial era because of the use of fossil fuels. The Earth mean temperature today is about 1 degree Celsius (C) higher than pre-industrial times. Recent scientific advances increasingly suggest that exceeding 1.5 degrees C may cause the Earth system to lurch through a cascading set of 'tipping points' - states of no return - driving an irreversible shift to a hotter world.

Climate change is global yet its manifestations and impacts will differ across the planet. Therefore, quantifying future changes at regional and local scales is critical for informed policy formulation. This, however, remains a significant challenge. We begin with a discussion on state-of-the-art science and technology on these questions and their current limitations. With this context, we come to the main theme of this article which is the potential of the emerging paradigm of 'quantum computing', and in particular Quantum Artificial Intelligence (QAI), in providing some of the breakthroughs necessary in climate science.

The rest of the chapter is organized as follows. In Sect. 2, we discuss science of climate change and the role of artificial intelligence. Section 3 discusses quantum artificial intelligence for the science of climate change. Section 4 concludes the chapter and highlights possible future directions.

## 2. Science of climate change and the role of artificial intelligence

Climate models have become indispensable to studying changes in the Earth's climate, including its future response to anthropogenic forcing. Climate modelling involves solving sets of coupled partial differential equations over the globe. Physical components of the Earth system - the atmosphere, ocean, land, cryosphere and biosphere - and the interactions between them are represented in these models and executed on high performance supercomputers running at speeds of petaflops and beyond. Models operate by dividing the globe into grids of a specified size, defined by the model resolution. The dynamical equations are then solved to obtain output fields averaged over the size of the grid. Therefore, only physical processes operating at spatial scales larger than the grid size are explicitly resolved by the models based on partial differential equations; processes that operate at finer scales, such as clouds and deep convection, are represented by approximate empirical relationships called parameterizations. This presents at least two significant challenges:

1. While climate models have become increasingly comprehensive, grid sizes of even state-of-the-art models are no smaller than about 25 km, placing limits on their utility towards *regional* climate projections and thereby for targeted policymaking.
2. Physical processes organizing at sub-grid scales often critically shape *regional* climate. Therefore, errors in their parameterizations are known to be the source of significant uncertainties and biases in climate models. Additionally, numerous biophysical processes are not yet well understood due to the complex and nonlinear nature of the interactions between the oceans, atmosphere and land.

Therefore, rapid advances are necessary to 'downscale' climate model projections to higher resolutions, improving parameterizations of sub-grid scale processes and quantifying as yet poorly understood non-linear feedbacks in the climate system.

A significant bottleneck in improving model resolution is the rapid increase in the necessary computational infrastructure such as memory, processing power and storage. For perspective, an *atmosphere-only* weather model with deep convection explicitly resolved was recently run in an experimental mode with a

1km grid size. The simulation used 960 compute nodes on SUMMIT, one of the fastest supercomputers in the world with a peak performance of nearly 150 petaflops, yet achieved a throughput of only one simulated week per day in simulating a four month period. A full-scale climate model, including coupled ocean, land, biosphere and cryosphere modules, must cumulatively simulate 1000s of years to perform comprehensive climate change studies. Towards surmounting the challenge of this massive scaling up in computing power, there have recently been calls for a push towards "exascale computing" (computing at exaflop speeds) in climate research. While the technology may be within reach, practical problems abound in terms of how many centres will be able to afford the necessary hardware and the nearly GW scale power requirements of exascale computing that will require dedicated power plants to enable it.

Similar bottlenecks exist for improving parameterizations of sub-grid scale processes. Satellite and ground based measurements have produced a deluge of observational data on key climate variables over the past few decades. However, these datasets are subject to several uncertainties such as data gaps, and errors arising during data acquisition, storage and transmission. The emerging challenge is to process and distil helpful information from this vast data deluge.

Towards overcoming these challenges to improving climate projections, we discuss recent advances in Artificial Intelligence that have enabled new insights into climate system processes. These techniques, however, are also subject to their own limitations. It is in this context that we discuss how Quantum Artificial Intelligence may help overcome those limitations and advance both higher resolution climate model projections and reduce their biases.

When machines learn decision-making or patterns from the data, they gain what is known as artificial intelligence (AI). Climate science has seen an explosion of datasets in the past three decades, particularly observational and simulation datasets. Artificial Intelligence (AI) has seen tremendous developments in the past decade and it is anticipated that its application to climate science will help improve the accuracy of future climate projections. Recent research has shown that the combination of computer vision and time-series models effectively models the dynamics of the Earth system ([10]). It is anticipated that advances in this direction would enable artificial intelligence to simulate the physics of clouds and rainfall processes and reduce uncertainties in the present systems [27]. In addition to helping augment the representation of natural systems in climate models by using the now available high quality data, AI has also been proposed for climate change mitigation applications (6]) Other areas where AI is playing a leading role are the technologies of carbon capture, building information systems, improved transportation systems, and the efficient management of waste, to name a few [6].

There are however limitations to the present deep learning models, for example, their inability to differentiate between causation and correlation. Moreover, Moore's law is expected to end by about 2025 as it bumps up against fundamental physical limits such as quantum tunneling. With the increasing demands of deep learning and other software paradigms, alternate hardware advancements are becoming necessary [15].

**3. Quantum Artificial Intelligence for the science of climate change**

Artificial intelligence algorithms suffer from two main problems: one is the availability of good quality data and the other is computational resources for processing big data at the scale of planet Earth. The impediments to the growth of AI based modelling can be understood from the way language models have developed in the past decade. In the early days of their success, developments were limited to computer vision, while natural language processing (NLP) lagged behind. Many researchers tried to use different algorithms for NLP problems but the only solution that broke ice was increasing the depth of the neural networks. Present day GPT, BERT and T5 models are evolved versions from that era. Maximizing gains from the rapid advances in artificial intelligence algorithms requires that they be complemented by hardware developments; quantum computing is an emerging field in this regard.

Quantum computers (QC) represent a conceptually different paradigm of information processing based on the laws of quantum physics [28]. The fundamental unit of information for a conventional / classical computer is the bit, which can exist in one of two states, usually denoted as 0 and 1. The fundamental unit of information for a quantum computer on the other hand is the "qubit", a two-level quantum system that can exist as a superposition of the 0 and 1 states, interpreted as being simultaneously in both states although with different probabilities. What distinguishes quantum from classical information processing is that multiple qubits can be prepared in states sharing strong "non-classical" interactions called "entanglement" that simultaneously sample a much wider informational space than the same number of bits, thereby enabling, in principle, massively parallel computation. This makes quantum computers far more efficiently scalable than their classical counterparts for certain classes of problems.

One trend of quantum computing is the race to demonstrate at least one problem that remains intractable to classical computers, but which can be practically solved by a quantum computer. Google coined this feat "quantum supremacy", and claimed, not without controversy, to achieve it with its 540qubit Sycamore chip [18]. A research team in China introduced Jiuzhang, a new light-based special-purpose quantum computer prototype, to demonstrate quantum advantage in 2020 [19]. The University of Science and Technology of China has successfully designed a 66-qubit programmable superconducting quantum processor, named ZuChongzhi [20]. IBM plans to have a practical quantum chip containing in excess of one thousand qubits by 2023 [21].

Artificial intelligence on quantum computers is known as quantum artificial intelligence and holds the promise of providing major breakthroughs in furthering the achievements of deep learning. NASA has Quantum Artificial Intelligence Laboratory (QuAIL) which aims to explore the opportunities where quantum computing and algorithms address machine learning problems arising in NASA's missions [24]. The JD AI research center announced that they have a 15-year research plan for quantum machine learning. Baidu's open-source machine learning framework Paddle has a subproject called paddle quantum, which provides libraries for building quantum neural networks [25]. However, for practical purposes, the integration of AI and quantum computing is still in its infant stage. The use of quantum neural networks is developing at a fast pace in the research labs, however, pragmatically useful integration is in its infant stages [22, 23]. The current challenges to industrial-scale QAI include how to prepare quantum datasets, how to design quantum machine learning algorithms, how to combine quantum and classical computations and identifying potential quantum advantage in learning tasks [26]. In the past 5 years, algorithms using quantum computing for neural networks have been developed ([3], [4]). Just as the open-source TensorFlow, PyTorch and other deep learning libraries stimulated the use of deep learning for various applications, we may anticipate that software such as TensorFlowQ (TFQ), QuantumFlow and others, already in development will stimulate advances in QAI

**3.1 Complex problems in Earth system science: Potential for QAI**

Quantum artificial intelligence can be used to learn intelligent models of earth system science bringing new insights into the science of climate change. Quantum AI (QAI) can play an essential role in designing climate change strategies based on improved, high-resolution scientific knowledge powered by QAI. Recent studies (for example, [13]) have attempted to develop physics schemes based on deep learning. However, these are largely proof-of-principle studies in nascent stages. Challenges such as the spherical nature of the data over Earth, complex and non-linear spatio-temporal dynamics and others exist in AI for improved climate models. Various techniques such as cubed spheres and tangent planes have been proposed to address the spatial errors arising out of sphericity. QAI can further develop advanced physical schemes using AI by incorporating high-resolution datasets, more extended training, and hyper parameter optimization. A necessary condition for quantum speedup of classical AI is that the task in question can be parallelized for training. Present libraries such as TensorFlow and PyTorch offer both data and model parallelism capabilities. They have also been released for quantum computers and need to be further developed for industrial scale quantum computers of the future.

**3.2 Technological Solutions for implementing QAI**

Figure 1 shows the technological solutions for implementing Quantum Artificial Intelligence for the climate science. The main technologies to empower quantum artificial intelligence for climate science includes the resources in the two GitHub repositories on Awesome Quantum Machine Learning (QML) [29,30], QML [31], Pennylane [32], QEML (Quantum Enhanced Machine Learning) [33], TensorFlowQ [34] and NetKet [35].

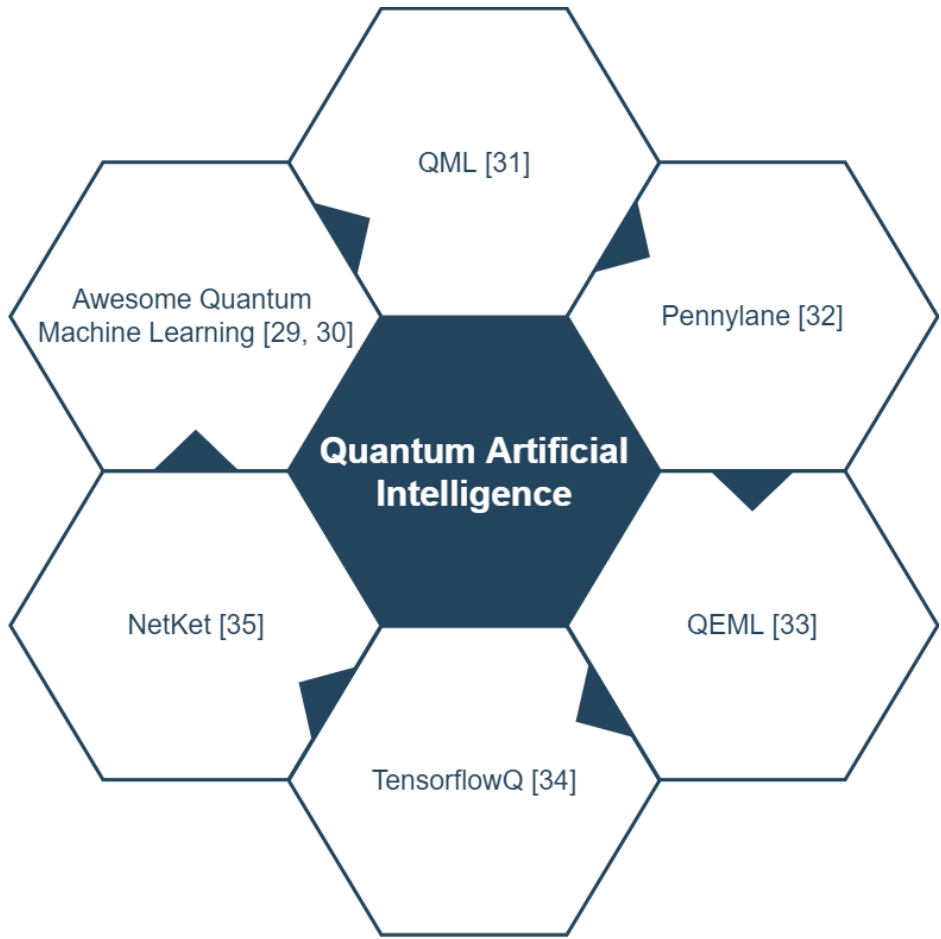

Figure 1: Technologies on Quantum artificial intelligence for the science of climate change

**3.3 Case study on the use of QAI for climate science**

We demonstrate an example of the application of Quantum Artificial Intelligence for land-use land cover classification on UC Merced dataset. The dataset is first transformed to Quantum data using Pennylane library and the training is

performed. The code can be found on the GitHub repository for this article at
https://github.com/manmeet3591/qai_science_of_climate_change.

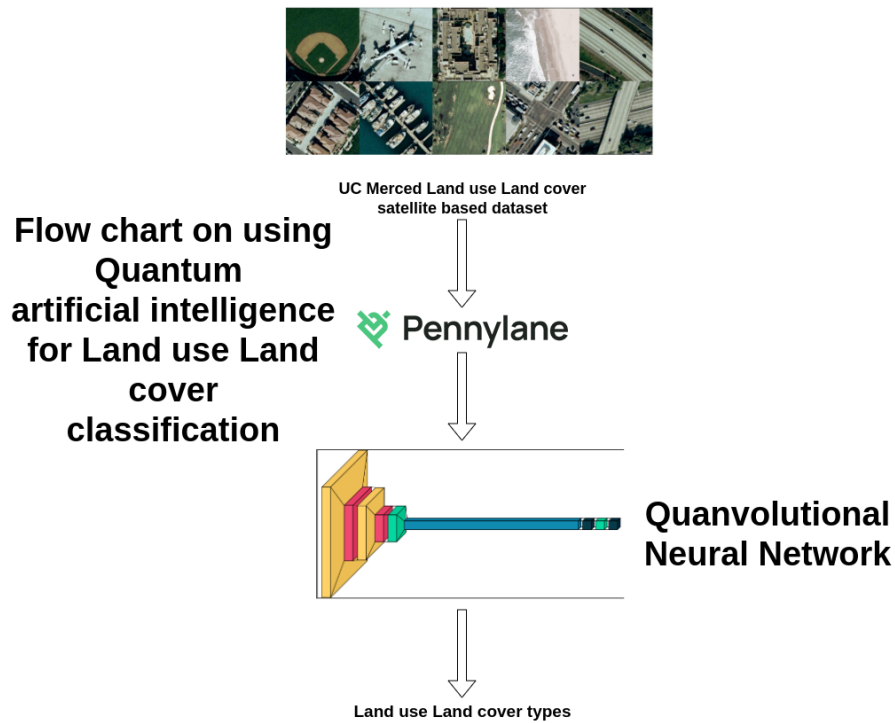

Figure 2: A case study on using Quantum artificial intelligence for land-use land-cover classification

Figure 2 shows the working of the case study. Initially, satellite data is transformed as Quantum data using Pennylane. Then, we apply the deep convolutional classification algorithm for land-use land-cover classification. The output classes consist of forests, agricultural fields etc

**4. Conclusions and Future Work**

Simulation studies are used to understand the science of climate change and are computationally expensive tools to understand the role of various forcings on the climate system. For example, recently, a study in the journal Science Advances showed how volcanic eruptions could force the coupling of El Nino Southern Oscillation and the South Asian Monsoon systems. Works of such kind are critical in advancing the understanding of the climate system and its response to various forcings. However, they are computationally demanding and are time-consuming to complete. Large ensemble climate simulations is an area that requires further work using Quantum computing and Quantum Artificial Intelligence. Quantum machine learning can play an important role, especially in pattern recognition for weather and climate science with problems that will benefit the most from quantum speedup being those that are inherently parallelizable.

However, various challenges present themselves in designing and operating useful quantum computers. State-of-the-art implementations of quantum computers today can control and manipulate on the order of 100 qubits whereas it is estimated that any real-world applications where quantum computers can reliably outperform classical computers would require on the order of a million qubits. This presents a formidable technological challenge. Additionally, entanglement - the heart of quantum computing - is a fragile resource prone to being destroyed with even the slightest disturbance (called "decoherence"). Therefore, operational quantum computers may be several years into the future. Yet, given their potential to affect a genuine paradigm shift, much effort has been invested in exploring its application to various fields and the focus of the present article concerns its possible climate science applications.

In summary, quantum artificial intelligence is projected to be a powerful technology of the future. Developments in the field of quantum artificial intelligence include the development of both computer vision and sequence algorithms capable of being implemented on large quantum computers. All these advancements would be driven by one factor, i.e. the development of high performance quantum computing hardware.

**Software Availability**

We have released the code for demonstrating the application of QAI on land-use land cover classification dataset from UC Merced dataset at: https://github.com/manmeet3591/qai_science_of_climate_change